\documentclass[conference]{IEEEtran} 
\usepackage{amsmath}
\usepackage{graphicx}
\usepackage{amssymb}
\usepackage{algorithm}
\usepackage{algpseudocode}

\title{Do We Need Transformers to Play FPS Video Games?}

\author{
\IEEEauthorblockN{Karmanbir Batth}
\IEEEauthorblockA{
    \textit{University of Waterloo} \\
    ksbatth@uwaterloo.ca
}

\and
\IEEEauthorblockN{Krish Sethi}
\IEEEauthorblockA{
    \textit{University of Waterloo} \\
    kasethi@uwaterloo.ca
}

\and
\IEEEauthorblockN{Aly Shariff}
\IEEEauthorblockA{
    \textit{University of Waterloo} \\
    ashariff@uwaterloo.ca
}

\and
\IEEEauthorblockN{Leo Shi}
\IEEEauthorblockA{
    \textit{University of Waterloo} \\
    l7shi@uwaterloo.ca
}

\and
\IEEEauthorblockN{Hetul Patel}
\IEEEauthorblockA{
    \textit{University of Waterloo} \\
    hr8patel@uwaterloo.ca
}
}

\begin{document}

\maketitle

\begin{abstract}
    
In this paper, we explore the Transformer based architectures for reinforcement learning in both online and offline settings within the Doom game environment. Our investigation focuses on two primary approaches: Deep Transformer Q-learning Networks (DTQN) for online learning \cite{reference7} and Decision Transformers (DT) for offline reinforcement learning \cite{reference10}. DTQN leverages the sequential modelling capabilities of Transformers to enhance Q-learning in partially observable environments,while Decision Transformers repurpose sequence modelling techniques to enable offline agents to learn from past trajectories without direct interaction with the environment. We conclude that while Transformers might have performed  well in Atari games, more traditional methods perform better than Transformer based method in both the settings in the VizDoom environment \cite{reference14}
\end{abstract}

\section{Introduction}

Q-networks traditionally have been used in various Atari game environments \cite{reference1} and in partially observable environments like in the Doom 1993 game \cite{reference2}. This was before the popularization of Transformers \cite{reference3}.

There is empirical evidence to show that Deep Transformer Q-networks outperform Deep Recurrent Q-networks in partially observable environments like memory cards and hallway \cite{reference7}. The model used to play Doom 1993 was a mix of DQN \cite{reference5} and DRQN \cite{reference6} models that were used for navigation and strategy \cite{reference2} respectively. We aim to see how Transformer can possibly increase the benchmarking results used in the paper \cite{reference2} which is the kill-death (k/d) ratio in a team deathmatch environment. This would be done by replacing the LSTM \cite{reference4} with the Transformer decoder architecture of the DTQN \cite{reference7}.

In recent years in Reinforcement Learning a new paradigm has emerged namely offline Reinforcement Learning \cite{reference10}.We evaluated the performance of offline reinforcement learning in VizDoom’s most basic scenario by benchmarking episode rewards, comparing the Decision Transformer architecture against the PPO model \cite{reference15}.

\subsection{Background}
VizDoom is a first-person shooter (FPS) environment that serves as a popular benchmark for Reinforcement Learning (RL) research. It is built upon Doom, a classic 1993 video game, and provides a highly customizable RL environment where agents interact with the environment using raw pixel-based observations.

Online reinforcement learning is a paradigm in which an agent interacts directly and continuously with an environment to learn optimal behaviors. The agent iteratively collects new experiences by actively taking actions and observing outcomes(states and rewards) from the environment to continuously update its policy.

Offline reinforcement learning is a paradigm that learns exclusively from static datasets of previously collected experiences \cite{reference11}. Offline RL applications primarily revolve around in catastrophic areas like robotics where trial and error while exploring policies can lead to catastrophic failures.
\subsection{Previous Work}
The combined usage of Deep Recurrent Q-Networks (DRQN) and DQN have demonstrated strong performance in partially observable environments, achieving optimal results in the VizDoom Deathmatch environment as presented in Playing FPS Games with Deep Reinforcement Learning \cite{reference2}. Similarly, Deep Transformer Q-Networks (DTQN) have shown their effectiveness in partially observable environments, where they achieved optimal results, highlighting the potential of transformer-based architectures for reinforcement learning in FPS video games like Doom \cite{reference7}. Although, they have not been used in VizDoom environment so far.

In the offline reinforcement learning domain, Decision Transformer (DT), originally proposed in \cite{reference10}, successfully learned optimal policies in MuJoCo environment, demonstrating the viability of sequence modeling approaches in reinforcement learning.

More recently, the RATE (Recurrent Action Transformer with Memory) paper \cite{reference12} demonstrated the successful training of a Decision Transformer (DT) in the ViZDoom TwoColours environment. In this scenario, the agent did not need to employ a strategy for shooting enemies but instead had to focus on strategic item collection and movement. This highlights that while transformer-based models can be effective in certain structured environments, their applicability in action-intensive tasks like shooting enemies remains an open challenge.

In both the settings, DT and DTQN have never been used in a VizDoom environments where the agent had to kill its enemies.
\begin{figure}[h]
    \centering
    \includegraphics[width=0.5\textwidth]{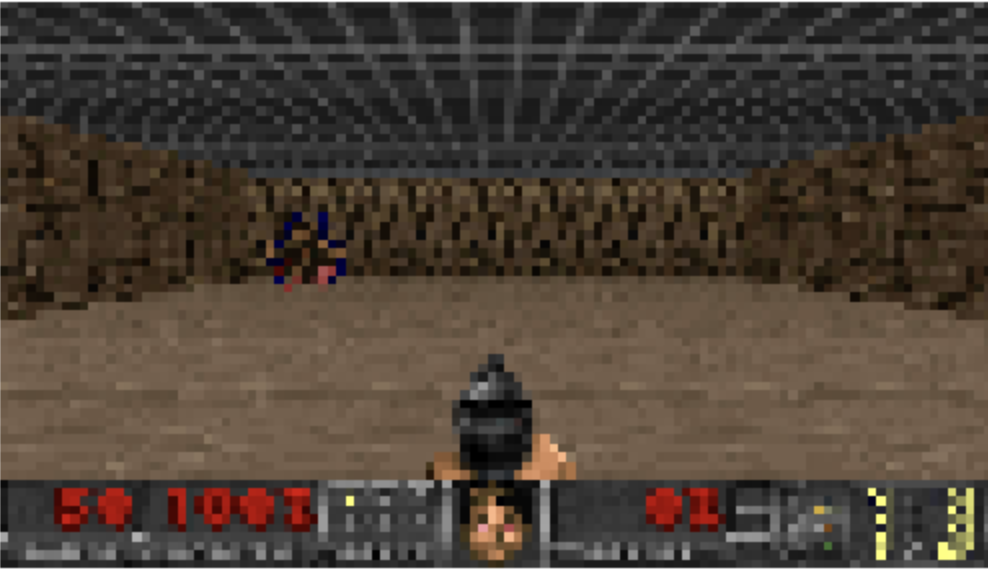}
    \caption{A picture of the agent playing the basic scenario in Doom}
    \label{}
\end{figure}

\section{Model Architecture}
\subsection{Deep Transformer Q-Networks}
A \textbf{Deep Transformer Q-Network (DTQN)} is an extension of \textbf{Deep Q-Networks (DQN)} that integrates transformer decoder based architecture modeled using Partially Observable Markov Decision Processes (POMDPs). Unlike conventional DQNs that use a fixed-size state representation, DTQN processes sequences of observations to learn a more context-aware representation of the environment. This allows DTQNs to maintain memory over long-horizon tasks, making them effective in environments with partial observability.

Formally, DTQN replaces the traditional fully connected layers in DQN with masked multi-head self-attention layers, enabling the network to capture temporal dependencies over sequences of observations.

\text{$Q$-Learning}'s goal is to learn a function \( Q : S \times A \to \mathbb{R} \), which represents the expected cumulative reward if starting in a state \( s \) and taking action \( a \). Specifically, Deep Q-Networks (DQN) are trained to minimize the Mean Squared Bellman Error \cite{reference1}
\[
\mathcal{L}(\theta) = \mathbb{E}_{(s,a,r,s') \sim \mathcal{D}} \left[ \left( r + \gamma \max_{a' \in \mathcal{A}} Q(s', a'; \theta') - Q(s, a; \theta) \right)^2 \right]
\]

\subsection{Descion Transformer}
\textbf{Decision Transformer (DT)} is a sequence modeling approach for reinforcement learning that uses \textbf{transformers} to predict actions based on past states, actions, and rewards. It reformulates reinforcement learning (RL) as a conditional sequence modeling problem, where a trajectory is treated as a sequence of tokens. Given a desired return-to-go (RTG), the model generates actions that maximize future rewards. The key advantages of Decision Transformers include offline reinforcement learning capabilities, scalability, and the ability to handle long-range dependencies through self-attention mechanisms. It was introduced as an autoregressive model trained on supervised learning objectives rather than traditional RL-based value function optimization.
\\Mathematically, a Decision Transformer models a trajectory as:
\[
\tau = (R_t, s_t, a_t, R_{t+1}, s_{t+1}, a_{t+1}, \dots)
\]
where \( R_t \) is the return-to-go, \( s_t \) is the state, and \( a_t \) is the action at timestep \( t \). The model predicts \( a_t \) based on previous tokens using a \textbf{causal transformer}.

Both Decision Transformers and Deep Transformer Q-Networks leverage the power of transformers in reinforcement learning, but DT focuses on offline RL with conditioned generation, while DTQN enhances Q-learning for partially observable environments.
\renewcommand{\thefigure}{2} 
\begin{figure}[h]
    \centering
    \includegraphics[width=0.5\textwidth]{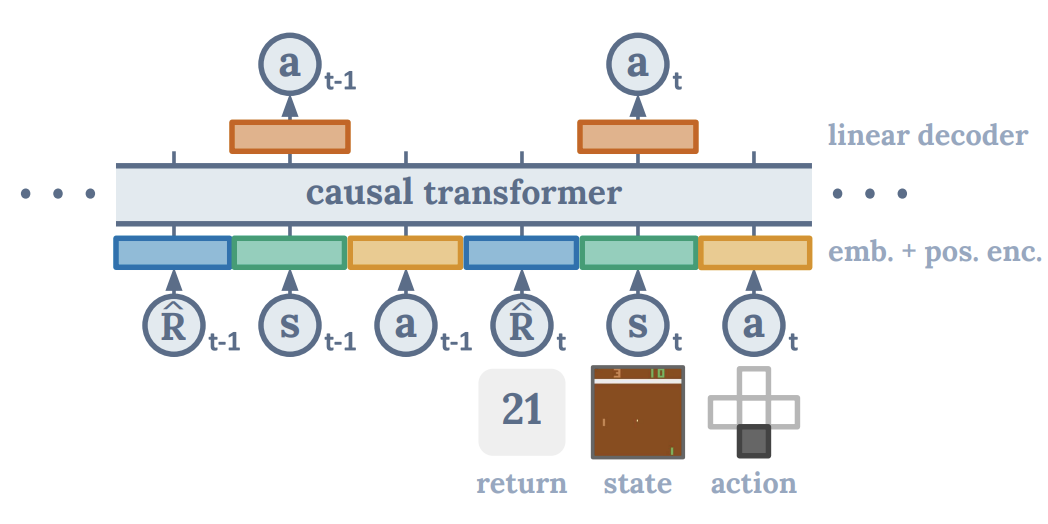}
    \caption{ is the DT architecture}
    \label{fig:DTQN_return}
\end{figure}

\renewcommand{\thefigure}{3} 
\begin{figure}[h]
    \centering
    \includegraphics[width=0.5\textwidth]{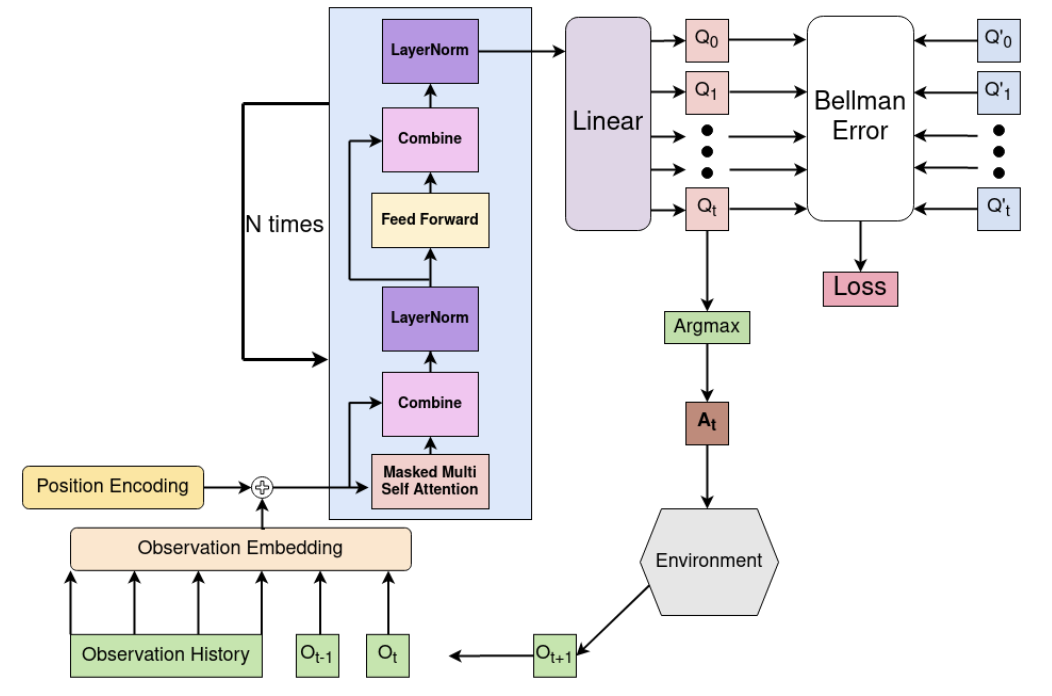}
    \caption{ is the DTQN architecture}
    \label{fig:DTQN_return}
\end{figure}

\section{Methodology}

\subsection{Deep Transformer Q-Networks}
The training loop for the DTQN follows a reinforcement learning approach to optimize the Q-function, \( Q(s, a; \theta) \), where \( \theta \) represents the network parameters. The details of the algorithm are adapted from the original DTQN paper \cite{reference7}. Here is an overview of the algorithm.
    




\textbf{1. Minibatch Sampling:} A minibatch of experience tuples \( (h_{t:t+k}, a_{t:t+k}, r_{t:t+k}, h_{t+1:t+k+1}) \) is sampled from the replay buffer \( \mathcal{D} \), containing sequences of states, actions, rewards, and future states.

\textbf{2. Target Q-value Calculation:} The target target Q-values for each time step \( t+i-1 \) is computed using the Bellman equation:
\[
r_{t+i-1} + \max_{a' \in \mathcal{A}} Q(h_{t+1:t+i+1}, a'; \theta')
\]
where \(r_{t+i-1}\) is the immediate reward at that timestep \( \theta' \) represents the target network, which is updated less frequently than the main network.

\textbf{3. Loss and Backpropagation:} The loss for each minibatch is the mean squared error between the predicted and target Q-values. The network parameters \( \theta \) are then updated using gradient descent. 

Our model processes visual input using a deep sequence modeling architecture. A set of 50 frames is fed to a CNN network that reduces the dimensionality of the images and encodes spatiotemporal features, producing a sequence of embeddings once flattened, which are augmented with sinusoidal positional encoding. The frames are then passed to 5 transformer layers that perform multi-head attention (8 attention heads), alongside residual gating. Finally, the results are passed to a feed-forward network to produce Q-values corresponding to each possible action. During training, actions are selected according to an \(\epsilon \)-greedy policy. The average training time for this setting was 20 hours, with 1.5 million iterations.

In addition to training a Q-network, our CNN layers are enhanced using a game features network during the training phase, which is used to predict features such as health, ammo, and number of enemies, for each given frame [5]. These features are only available during training time, which we leveraged to improve the embeddings for the transformer.

Our reward shaping follows the method explained in [5].

\subsection{Decision Transformer }
The training methodology for the Decision Transformer (DT) closely follows the approach outlined by the work done in recurrent action transformer with memory \textit{Recurrent Action Transformer with Memory} \cite{reference12}. We adopted their hyperparameters, except for the targeted return, which we set to 110 since it gave the most optimum result. Decision Transformers are trained using a dataset of past trajectories collected from an environment. Each trajectory consists of a sequence of states, actions, and rewards:

\begin{equation}
    \tau = \{(s_1, a_1, r_1), (s_2, a_2, r_2), \dots, (s_T, a_T, r_T)\}
\end{equation}

During training, the transformer model processes a context length of 90 past trajectories and learns to predict the next action \( a_t \) given the current state \( s_t \) and returns-to-go (RTG), which represents the expected future reward:

\begin{equation}
    a_t = \pi_{\theta}(s_t, r_t, \text{RTG}_t)
\end{equation}

where RTG is computed as:

\begin{equation}
    \text{RTG}_t = \sum_{k=t}^{T} \gamma^{k-t} r_k
\end{equation}

\noindent where \( \gamma \) is the discount factor.

We generated a dataset of 45,000 trajectories using a PPO model implemented with the Stable-Baselines3 python library \cite{reference15}. The environment was based on ViZDoom, and data collection began after 90,000 training steps of the PPO model. During training, the PPO model achieved an average episode return of approximately 80, with returns ranging from -400 to 100. 

Training for the DT was conducted over 100 epochs. To preprocess observations, we employed convolutional neural networks (CNNs) with hyperparameters sourced from Deep Reinforcement Learning on a Budget: 3D Control and Reasoning Without a Supercomputer \cite{reference13}. The Decision Transformer was then trained on the preprocessed observations extracted from the collected dataset. 

The environment for training DT was the \textbf{basic scenario} in the ViZDoom library, using the default reward structure, action space, and other configuration settings provided by ViZDoom.

To improve training efficiency, we employed the \textbf{frame skipping technique} with a frame skip of 4 \cite{reference2}. This means that the agent repeated the same action for the next four frames without interacting with the environment. Consequently, the agent received a new screen input only every \( k + 1 \) frames, where \( k \) is the number of frames skipped between steps (in our case it was 4). The average training time was 45 mins.

\section{Results}
To evaluate the effectiveness of transformer-based architectures in reinforcement learning, we conducted experiments in both offline and online RL settings.

For the \textbf{online RL} setting, we compared the Deep Transformer Q-Network (DTQN) against the combined usage of DQN and  Deep Recurrent Q-Network (DRQN) wich used the LSTM model. Our findings demonstrate that DQN-DRQN consistently achieved better performance in VizDoom's Team Deatchmatch setting, which involved different sets of maps in DOOM video game. DTQN's reliance on self-attention did not effectively compensate for missing state information, whereas DQN-DRQN's explicit recurrence allowed for better state tracking and decision-making under uncertainty.

In the \textbf{offline RL} setting, we compared the episode rewards for Decision Transformer (DT) and Proximal Policy Optimization (PPO) in VizDoom's basic scenario. Results indicate that PPO outperformed the Decision Transformer in terms of final policy performance.

These results suggest that while transformers offer strong sequence modeling capabilities, they are not inherently well-suited for environments requiring active memory-based strategies like VizDoom. In both the settings transformer-based architectures learnt a suboptimum policy, while more traditional methods learnt a better policy 
\\
\includegraphics[width=0.5\textwidth]{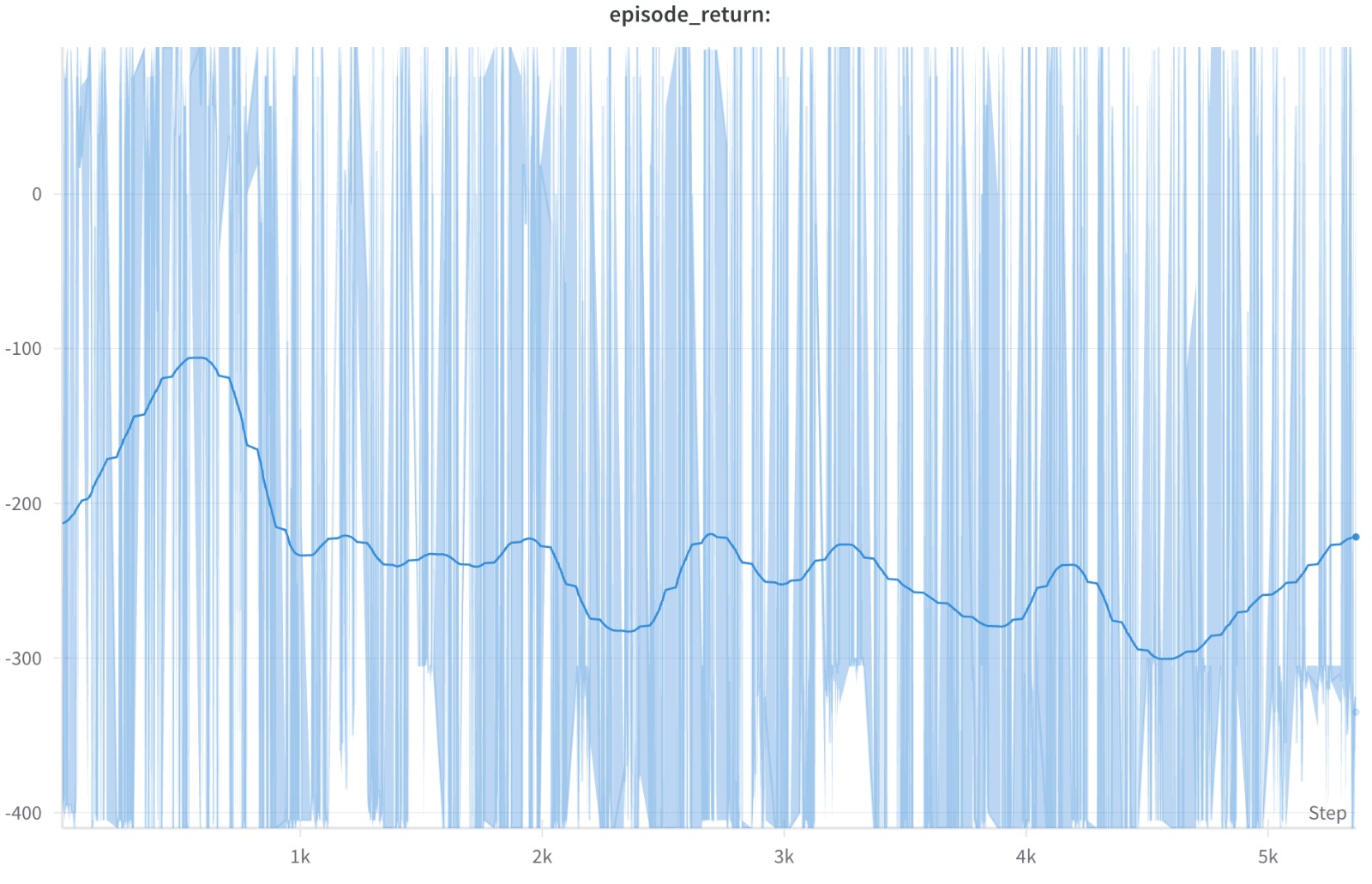} 
Fig 4: the episode returns of the DT architecture
\includegraphics[width=0.5\textwidth]{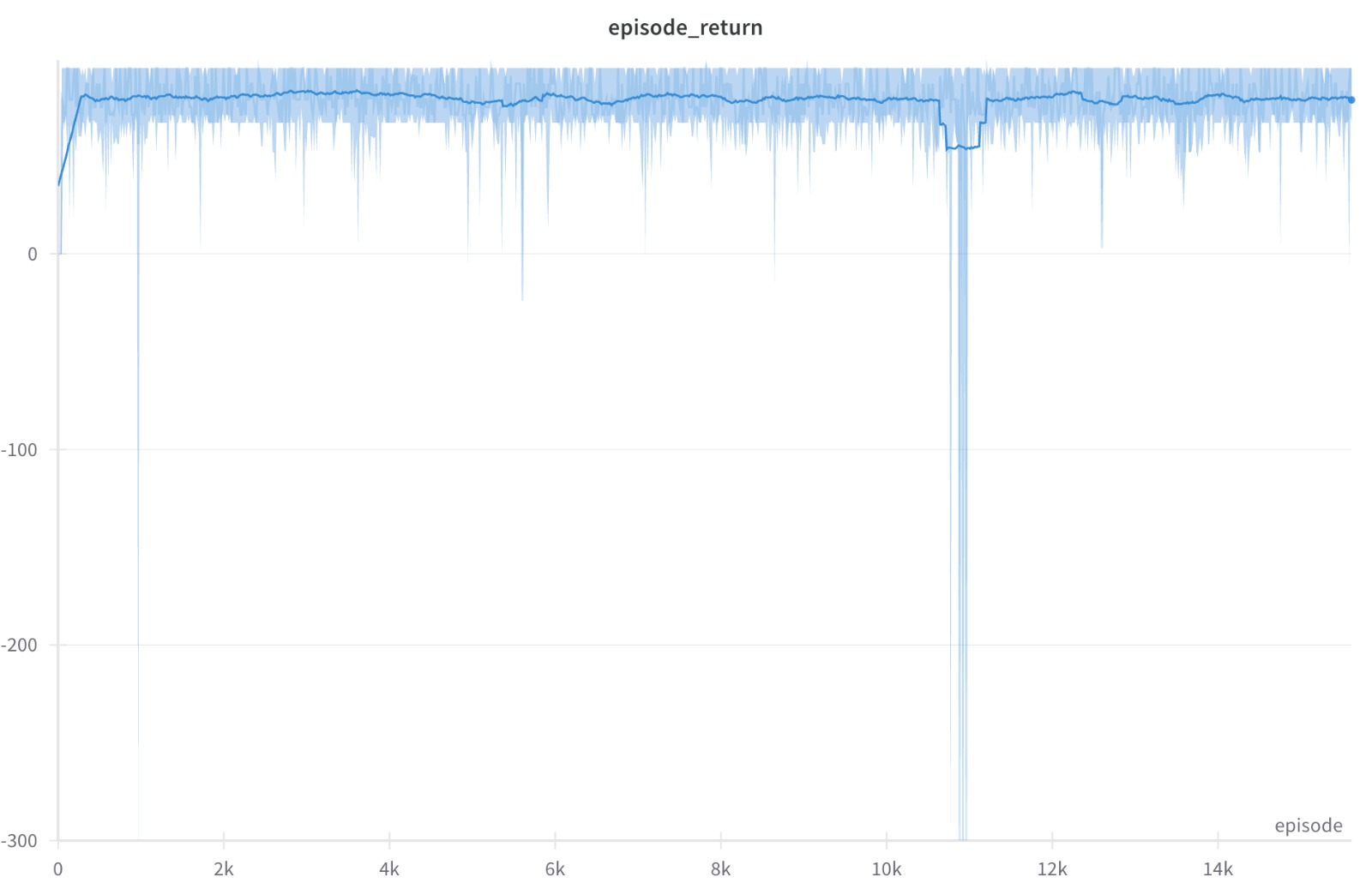} 
  Fig 5:  the episode returns of the PPO model.
\\
\renewcommand{\thefigure}{6} 
\begin{figure}[h]
    \centering
    \includegraphics[width=0.5\textwidth]{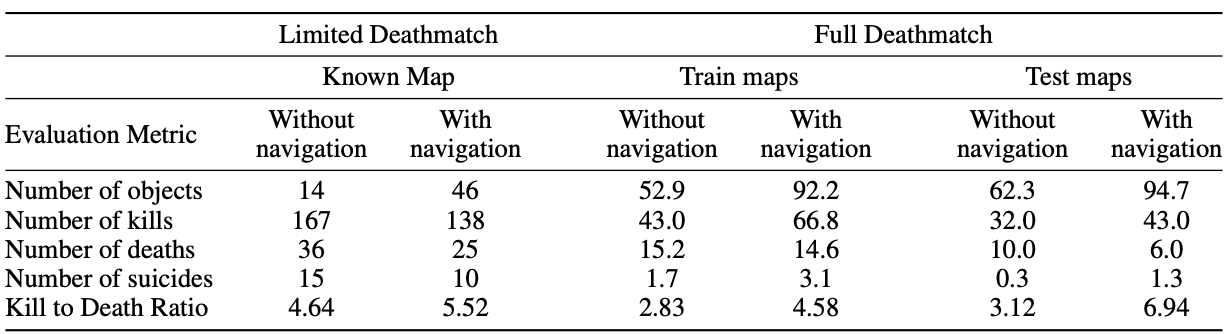}
    \caption{ the k/d ratio achieved by the DQN-DRQN model \cite{reference2}}
    \label{fig:DTQN_return}
\end{figure}
\renewcommand{\thefigure}{\arabic{figure}} 

\renewcommand{\thefigure}{7} 
\begin{figure}[h]
    \centering
    \includegraphics[width=0.5\textwidth]{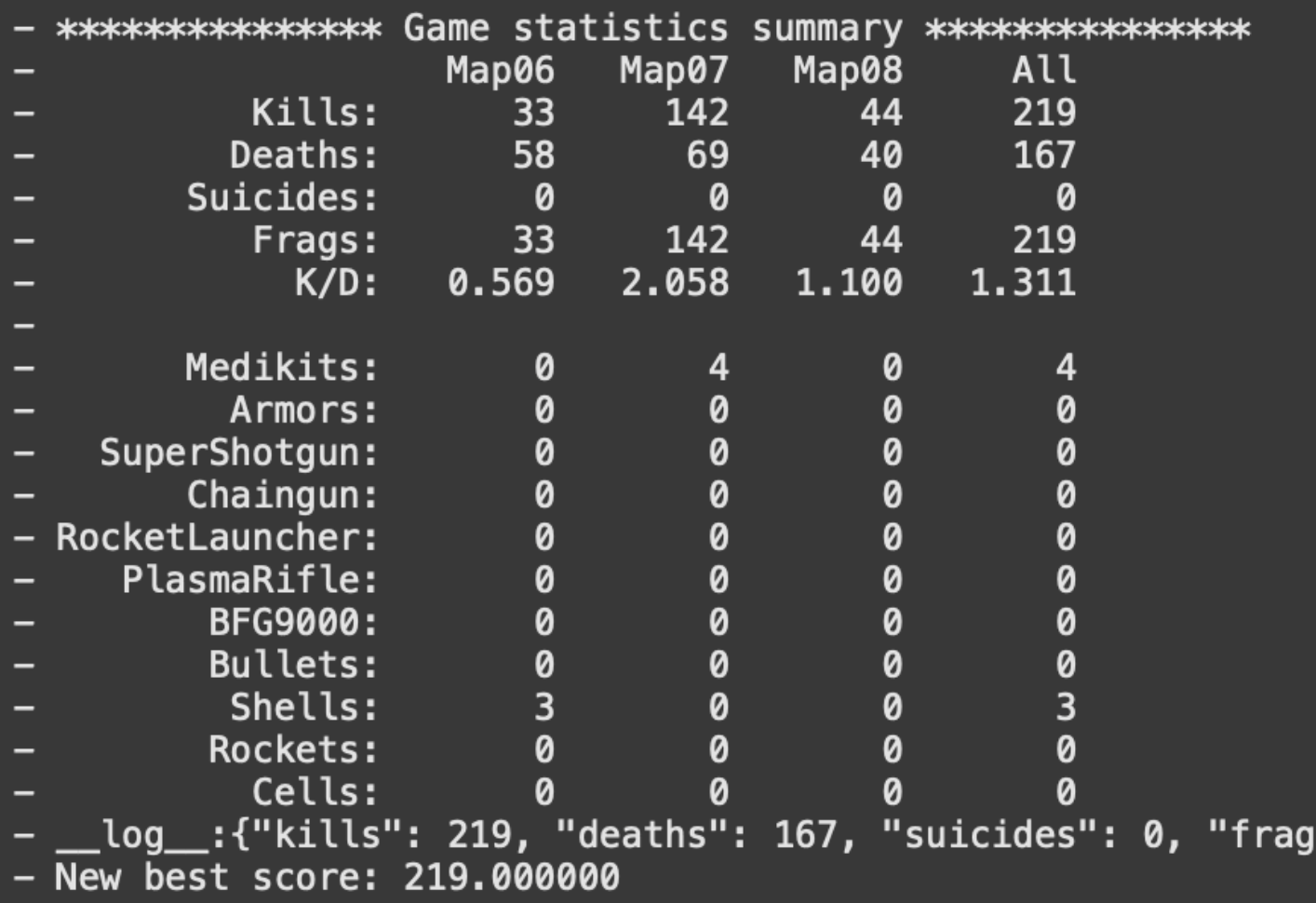}
    \caption{DTQN k/d ratio}
    \label{fig:DTQN_return}
\end{figure}
\renewcommand{\thefigure}{\arabic{figure}} 

\section{Conclusion}
In this work, we investigated the application of transformer-based architectures in reinforcement learning for FPS video games using the VizDoom environment. We evaluated two primary models: Deep Transformer Q-Networks (DTQN) for online reinforcement learning and Decision Transformers (DT) for offline reinforcement learning. Our goal was to assess whether transformer-based architectures outperform traditional methods in highly memory and strategy intensive environments like Doom.

Our results indicate that while transformers provide strong sequence modeling capabilities, they struggle in highly partial observable settings requiring strategic decision-making. In the \textbf{online RL} setting, DTQN underperformed compared to the combined usage of Deep Q-Networks (DQN) and Deep Recurrent Q-Networks (DRQN) with LSTMs. The recurrence in DRQN enabled more effective state tracking and decision-making, while DTQN's reliance on self-attention alone was insufficient to compensate for missing state information.

In the \textbf{offline RL} setting, we compared Decision Transformers (DT) against Proximal Policy Optimization (PPO) using a dataset of 45,000 trajectories. The empirical results show that PPO achieved superior policy performance in VizDoom’s basic scenario. The Decision Transformer was able to learn meaningful policies from offline data, but struggled to generalize optimally, highlighting the limitations of sequence modeling approaches in memory intensive environments requiring strategy.

Overall, our findings suggest that while transformer-based architectures have demonstrated success in certain reinforcement learning benchmarks, they are not inherently well-suited for FPS environments (specifically VizDoom) that are memory intensive environments requiring strategy to move and kill enemies. The reliance on self-attention alone may not be sufficient to capture long-term dependencies in these settings. 

A potential direction for future research is to investigate architectures that do not rely on self-attention but still perform well on long-range sequence tasks. One promising approach is Decision Mamba \cite{reference9}, which replaces self-attention with selective state-space models. This could provide an efficient way to handle long-range strategy while maintaining scalability in reinforcement learning environments.

Overall, while Transformer-based models have demonstrated success in structured tasks such as Atari \cite{reference1} and some partially observable environments like hallway \cite{reference7}, their application in first-person shooter (FPS) reinforcement learning remains a challenge, requiring further research in attention mechanisms and memory modeling.
\pagebreak

\bibliographystyle{IEEEtran}
\bibliography{references}
\end{document}